\documentclass{article}
\usepackage{spconf}
\usepackage[reqno]{amsmath}
\usepackage{amssymb}
\usepackage{amsthm}
\usepackage{mathtools}
\usepackage{graphicx}
\usepackage{physics}
\usepackage{upgreek}
\usepackage{anyfontsize}
\usepackage{tabularray}
\UseTblrLibrary{booktabs,siunitx}
\usepackage{capt-of}
\usepackage{duckuments}
\usepackage[dvipsnames]{xcolor}
\usepackage{confnames}

\makeatletter
\renewcommand{\section}{\@startsection{section}{1}{\z@}%
                                   {-3.5ex \@plus -1ex \@minus -.2ex}%
                                   {2.3ex \@plus.2ex}%
                                   {\centering\normalfont\normalsize\bfseries\uppercase}}
\makeatother

\DeclareSIUnit[quantity-product=]\percent{\%}
\DeclareMathSizes{10}{10}{6}{6}
\newcommand\defeq{\stackrel{\mathclap{\fontsize{4pt}{4pt}\selectfont\text{def}}}{=}}

\newif\ifsubmit
\submitfalse

\allowdisplaybreaks[4]

\graphicspath{{./figures}}

\usepackage{CJKutf8}

\definecolor{LightOrange}{rgb}{0.98,0.9,0.84}

\newcommand{\tablefontsize}{\fontsize{8.5pt}{9pt}\selectfont}
\newcommand{\DA}{\downarrow}
\newcommand{\UA}{\uparrow}
\NewTableCommand{\BB}{\SetCell{font=\bfseries}}

\usepackage[colorlinks]{hyperref}
\usepackage[capitalize]{cleveref}

\newtheorem{proposition}{Proposition}
\crefname{proposition}{Proposition}{Prop.}
\crefname{section}{Sec.}{Sec.}

\newcommand{\projpage}{\url{https://github.com/0V/ESIM-AD.git}}

\allowdisplaybreaks

\newcommand{\heading}[1]{\vspace{0.3em}\noindent\textbf{#1:}\,}

\title{Event-based Camera Simulation using Monte Carlo Path Tracing\\with Adaptive Denoising}
\name{Yuta Tsuji$^1$ ~~~ Tatsuya Yatagawa$^{2,\dagger}$ ~~~ Hiroyuki Kubo$^3$ ~~~ Shigeo Morishima$^1$\thanks{$^\dagger$ Corresponding author. Email: \href{mailto:tatsuya.yatagawa@r.hit-u.ac.jp}{tatsuya.yatagawa@r.hit-u.ac.jp}}}
\address{$^1$Waseda University ~~~ $^2$Hitotsubashi University ~~~ $^3$Chiba University}

\begin{document}
%
\maketitle
\begin{abstract}
  This paper presents an algorithm to obtain an event-based video from noisy frames given by physics-based Monte Carlo path tracing over a synthetic 3D scene. Given the nature of dynamic vision sensor (DVS), rendering event-based video can be viewed as a process of detecting the changes from noisy brightness values. We extend a denoising method based on a weighted local regression (WLR) to detect the brightness changes rather than applying denoising to every pixel. Specifically, we derive a threshold to determine the likelihood of event occurrence and reduce the number of times to perform the regression. Our method is robust to noisy video frames obtained from a few path-traced samples. Despite its efficiency, our method performs comparably to or even better than an approach that exhaustively denoises every frame. Visit our project page for more information: \projpage.
\end{abstract}
\section{Introduction}
\label{sec:intro}

Event-based cameras equipped with a dynamic vision sensor (DVS) is a special camera device to detect brightness changes as ``events''~\cite{patrick2008latency}. Event-based cameras can detect events asynchronously and independently at each pixel, and the temporal resolution is significantly higher than the ordinary image sensors. These properties of event-based cameras/videos have been leveraged in many applications, such as object tracking~\cite{mitrokhin2018event}, 3D scanning~\cite{matsuda2015mc3d}, optical flow estimation~\cite{benosman2014event,Zhu2018evflownet,tian2022event}, and simultaneous localization and mapping (SLAM)~\cite{kim2016real,vidal2018ultimate}. For more information, refer to the comprehensive survey~\cite{gallego2022event}.

The goal of this study is to facilitate the above applications of event-based cameras by rendering event-based videos based on physics-based Monte Carlo (MC) path tracing. As we briefly review the previous studies below, the path-tracing-based rendering of event-based videos is still a challenging problem mainly due to its high computational complexity.

\subsection{Event-based cameras}
\label{ssec:event-cameras}

In the noise-free scenario, an event $e_{kl}$ is triggered at pixel $\vb{q}_k$ and at time $t_l$ when the brightness increment reaches a temporal contrast threshold $C \geq 0$. Assume that the last event is detected $\Delta t$ earlier than $t_l$, then an event is triggered when the following equation holds.
\begin{gather}
  \Delta L(\vb{q}_k, t_l) = L(\vb{q}_k, t_l) - L(\vb{q}_k, t_l - \Delta t) = p_k C,
  \label{eq:event-detect}
\end{gather}
where $L(\vb{q}_k, t_l)$ is the brightness at a spatio-temporal position $(\vb{q}_k, t_l)$, and $p_k \in \{ -1, 1 \}$ is a polarization of the event.The brightness $L$ is the logarithm of the incident irradiance $E$ on a sensor, i.e., $L = \log E$, which is inspired by the high sensitivity of the human visual system to darker components.

\subsection{Related work}
\label{ssec:related-work}

When we develop or test the applications, it will be useful to obtain the event-based video from RGB video frames with a common frame rate (e.g., 30--60 frames per second). To this end, previous studies have solved two problems, i.e., increasing the frame rate of an input video and transforming the video frames into a set of events. Based on the behavior of DVS sensors in \cref{eq:event-detect}, a pioneering work by Katz et al.~\cite{katz2012live} used a \SI{200}{\Hz} frame rate camera and synthesized DVS events with a \SI{5}{\ms} time resolution. Rebecq et al. proposed a system called ``ESIM''~\cite{rebecq2018esim}, which obtains events at adaptive time points from synthetic 3D scenes using a black-box rendering engine. However, when applying MC path tracing to the rendering engine, these systems would be impractical because obtaining noise-free video frames with MC path tracing requires considerable computation time. Recently, Hu et al. proposed another system called ``v2e''~\cite{hu2021v2e}, where the input video frames are increased in frame rate using a deep-learning-based method, i.e., Super SloMo~\cite{jiang2018super}, and then the high frame rate video is converted into a set of events by carefully considering the behavior of real DVS sensors. However, the v2e system also requires noise-free video frames, and we need significant computation time when we input video obtained by MC path tracing to solve the rendering equation~\cite{kajiya1986rendering}. Thus, efficient simulation of the event-based camera through MC path tracing is still challenging.

\section{Efficient Event-based Video Rendering}
\label{sec:method}

Our method leverages WLR-based image denoising~\cite{moon2014adaptive} but reduces the computation by solving the regression problem only at the pixels where an event is more likely to be detected. For this purpose, we theoretically derive a threshold to determine the likelihood of event occurrence.

\subsection{Background: weighted local regression for denoising}
\label{ssec:wlr}

\begin{figure}[t!]
  \centering
  \includegraphics[width=0.95\linewidth]{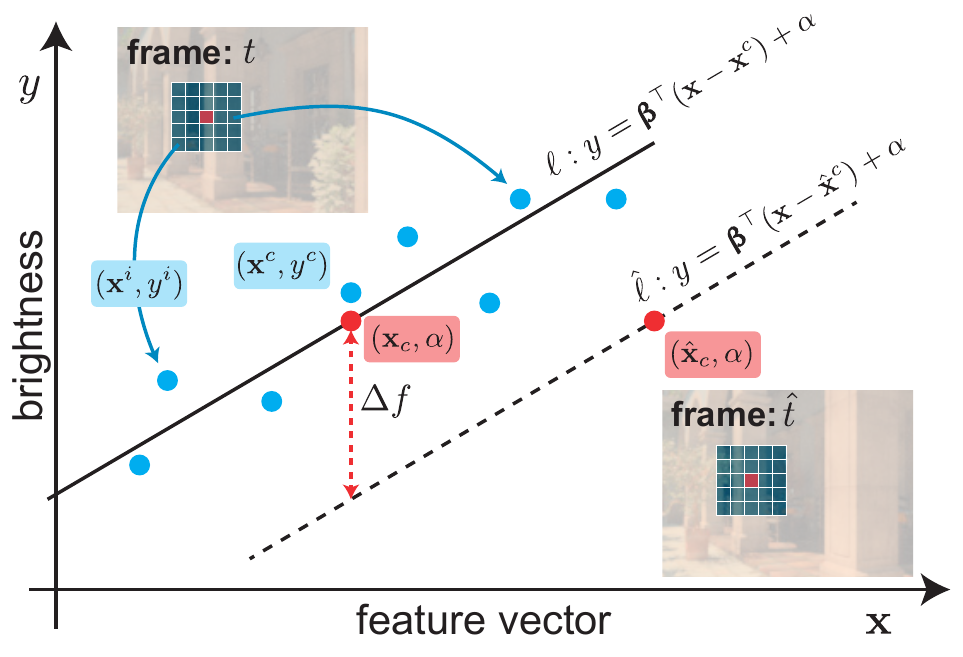}
  \caption{An illustration for our event detection method using weighted local regression. For a frame where the last event occurred, we regress a hyper-plane (i.e., a straight line in this figure) for a set of features $\{ \vb{x}^i \}$ centered by $\vb{x}^c$. Then, for a new frame, we translate the hyper-plane to go across a new centroid feature $\hat{\vb{x}}^c$ and calculate the weighted mean squared distance between the hyper-plane and the feature vectors at the time of the last event.}
  \label{fig:regression}
\end{figure}

Rather than collecting huge path samples to get noise-free video frames, we can denoise each noisy video frame obtained by a fewer number of path samples. Then, we can detect brightness changes using the denoised frames. Before introducing our method, we start with introducing an image denoising method~\cite{moon2014adaptive} based on the WLR. Assume a parametric curve or surface $f(\vb{x})$ is a function of feature vector $\vb{x}$. Then, its underlying stochastic model is
\begin{equation}
  y = f(\vb{x}) + \epsilon,
  \label{eq:stochastic-model}
\end{equation}
where $y \in \mathbb{R}$ is an observation including stochastic noise $\epsilon$. Our problem supposes $y \defeq L = \log E$ as denoted in the original paper~\cite{moon2014adaptive}. In this model, the unknown intensity $f(\vb{x})$ can be approximated locally based on the first-order Taylor expansion of \cref{eq:stochastic-model} around the central feature vector $\vb{x}^c$:
\begin{equation}
  f(\vb{x}) \approx f(\vb{x}^c) + \nabla f(\vb{x}^c)^\top (\vb{x} - \vb{x}^c).
  \label{eq:taylor}
\end{equation}
The feature vector $\vb{x}$ here is composed of world space position, normal vector, and texture color at the position that a light ray from the camera first hits~\cite{moon2014adaptive}. Here, $\alpha \defeq f(\vb{x}^c) \in \mathbb{R}$ and $\pmb{\upbeta} \defeq \nabla f(\vb{x}^c) \in \mathbb{R}^D$ are the unknowns of the model and regressed by minimizing the residual $\mathcal{R}$:
\begin{gather}
  \min_{\alpha, \pmb{\upbeta}} \mathcal{R} = \min_{\alpha, \pmb{\upbeta}} \frac{1}{W} \sum_{i \in \Omega_{c}} w^i (y^i - \alpha - \pmb{\upbeta}^\top (\vb{x}^i - \vb{x}^c))^2, \label{eq:minimization} \\
  w^i = \textstyle\prod_{j=1}^D w \qty( \frac{\vb{x}_j^i - \vb{x}_j^c}{h \vb{b}_j} ), \quad W = \sum_{i \in \Omega_{c}} w^i, \label{eq:weight-sum}
\end{gather}
where $w$ is a Gaussian fall-off function, $h$ and $\vb{b}$ are the parameters to control the magnitude of a fall-off radius for each dimension of the feature vectors. As illustrated in \cref{fig:regression}, the weighted sum is calculated with neighboring pixels indexed by $i$ of the local window $\Omega_c$ around the pixel $c$.

To denoise an image, we need to solve the minimization in \cref{eq:minimization} at each pixel. Although this minimization can be solved as a simple linear system with a $D \times D$ coefficient matrix, solving it for every pixel of a high frame rate video is impractically time-consuming.

\subsection{Reduced event detection by weighted local regression}
\label{ssec:wlr-event}

To reduce the number of times to solve the WLR, we derive a threshold to determine whether to solve it. Using \cref{eq:taylor}, we can represent the difference $\Delta f$ of the underlying noise-free brightness values $f(\vb{x})$ of two different frames as
\begin{align}
  \Delta f = \abs{ f(\vb{x}^c) - f(\hat{\vb{x}}^c) }
   &\approx \left| (f(\vb{x}^c) + \nabla f(\vb{x}^c)^\top (\vb{x}^c - \vb{x}^c)) \right. \notag \\
   &\quad - \left. (f(\vb{x}^c) + \nabla f(\vb{x}^c)^\top (\hat{\vb{x}}^c - \vb{x}^c)) \right| \notag \\
   &= | \pmb{\upbeta}^\top (\vb{x}^c - \hat{\vb{x}}^c) |.
  \label{eq:abs-brightness}
\end{align}
As illustrated in \cref{fig:regression}, this represents the distance of two parallel lines $\ell: y = \pmb{\upbeta}(\vb{x} - \vb{x}^c) + \alpha$ and $\hat{\ell}: y = \pmb{\upbeta}(\vb{x} - \hat{\vb{x}}^c) + \alpha$. Therefore, calculating \cref{eq:abs-brightness} and threshold it with $C$ will be a simple solution. However, as we mentioned previously, the feature vector $\vb{x}$ includes the noisy brightness of a pixel obtained by a few path samples. Therefore, the right-hand side of \cref{eq:abs-brightness} must include noise. To alleviate this problem, we instead compute the following residual for event detection.
\begin{equation}
  \hat{\mathcal{R}} = \frac{1}{W} \sum_{i \in \Omega_c} w^i (y^i - \alpha - \pmb{\upbeta}^\top (\vb{x}^i - \hat{\vb{x}}^c))^2,
  \label{eq:new-residue}
\end{equation}
which means a weighted average of the distance $\hat{\ell}$ and the neighboring features of $\vb{x}^c$ (not $\hat{\vb{x}}^c$). For \cref{eq:new-residue}, the following proposition about thresholding holds.

\begin{proposition}
  \label{prop:threshold}
  Assume that an event is detected when $\Delta L = p C$. Then the event can be detected by thresholding the difference of the residues $\Delta \mathcal{R} = \abs{ \hat{\mathcal{R}} - \mathcal{R} }$ with $C^2$.
\end{proposition}

\begin{proof}
  We can write $\hat{\mathcal{R}}$ using $\mathcal{R}$ as follows:
  \begin{align}
    \hat{\mathcal{R}} \!
     &= \frac{1}{W} \sum_{i \in \Omega_c} w^i (y^i - \alpha - \pmb{\upbeta}^\top (\vb{x}^i - \vb{x}^c) - \pmb{\upbeta}^\top (\vb{x}^c - \hat{\vb{x}}^c))^2 \notag \\
     &= \mathcal{R} + \qty( \pmb{\upbeta}^\top (\vb{x}^c - \hat{\vb{x}}^c) )^2 \sum_{i=1}^n \frac{w^i}{W} \notag \\
     &\quad -2 \pmb{\upbeta}^\top (\vb{x}^c \!- \hat{\vb{x}}^c) \sum_{i=1}^n \frac{w^i}{W} (y^i \! - \alpha - \pmb{\upbeta}^\top (\vb{x}^i \!- \vb{x}^c)). \label{eq:new-residue-simplify}
  \end{align}
  Because $\alpha$ and $\pmb{\upbeta}$ minimize $\mathcal{R}$, the derivative of $\mathcal{R}$ with respect to $\alpha$ is approximately zero; hence, the last term of \cref{eq:new-residue-simplify} is also approximately zero. Also, the sum $\frac{1}{W} \sum_i w^i = 1$ according to \cref{eq:weight-sum}. Therefore, we have
  \begin{equation}
    \Delta \mathcal{R} \approx \qty( \pmb{\upbeta}^\top (\vb{x}^c - \hat{\vb{x}}^c) )^2.
    \label{eq:delta-residue}
  \end{equation}
  Considering \cref{eq:abs-brightness}, \cref{eq:delta-residue}, and $\abs{ f(\vb{x}^c) - f(\hat{\vb{x}}^c) } = \abs{p C} = C$ when an event occurs, we can use $C^2$ to threshold $\Delta \mathcal{R}$.
\end{proof}

\noindent
\Cref{prop:threshold} says that $\Delta \mathcal{R}$ and $\pmb{\beta}^\top (\vb{x} - \hat{\vb{x}}^c)$ are approximately equal. However, feature vectors, such as $\vb{x}^c$ and $\hat{\vb{x}}^c$, include noise, and therefore, evaluating $\Delta \mathcal{R}$ for detecting events is more robust to noise. This is the reason that we do not directly evaluate $\pmb{\upbeta}^\top(\vb{x}^c - \hat{\vb{x}}^c)$, even though this can also be used for the same purpose and is computationally simpler.

\section{Experiments}
\label{sec:experiments}

\begin{figure}[t!]
  \centering
  \includegraphics[width=\linewidth]{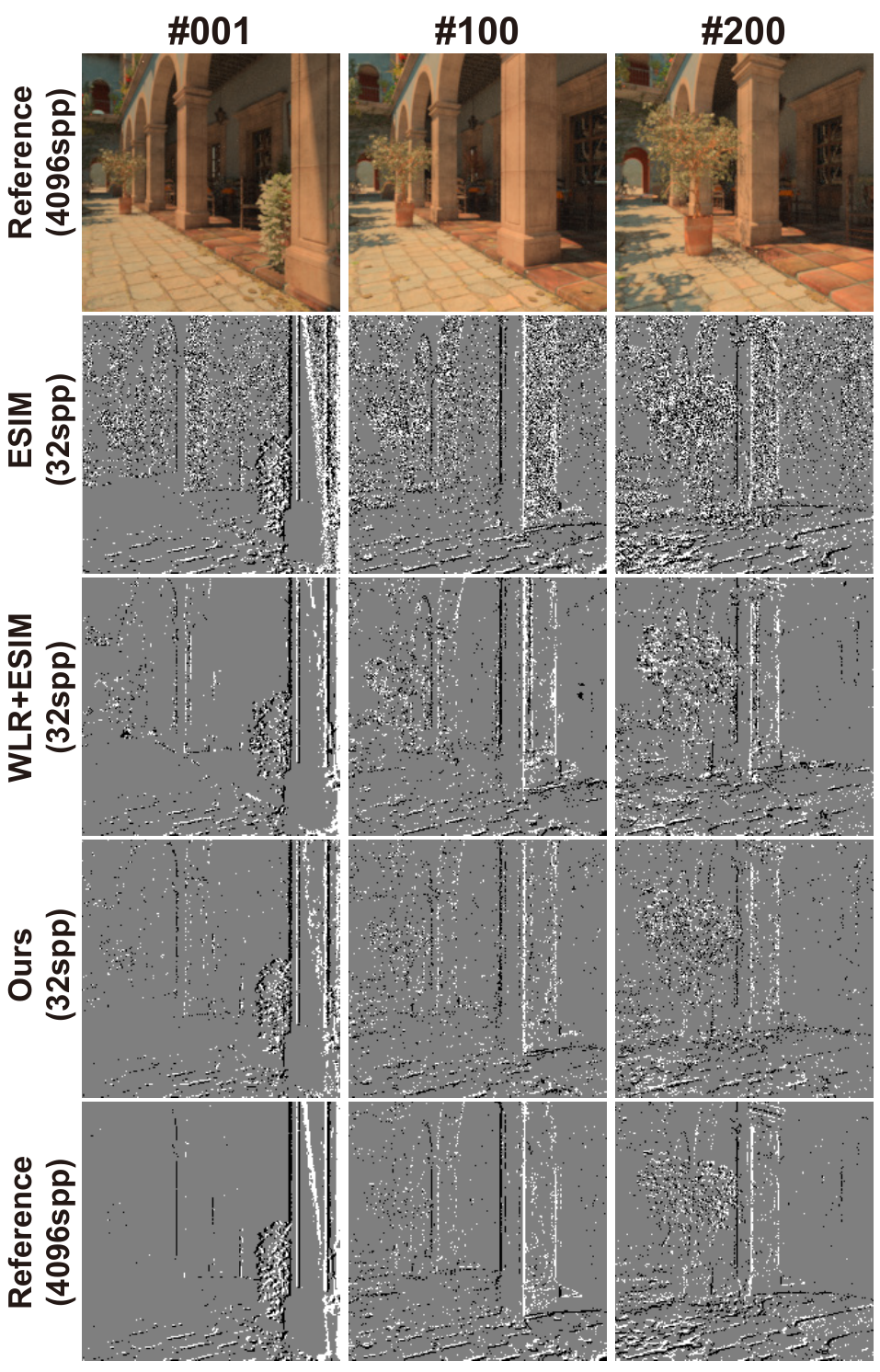}
  \caption{Visual comparison of event-based video frames. In addition to the 1st, 100th, and 200th frames shown here, the full-length videos are available on our project page.}
  \label{fig:results}
\end{figure}

\begin{figure*}[t!]
  \centering
  \includegraphics[width=\linewidth]{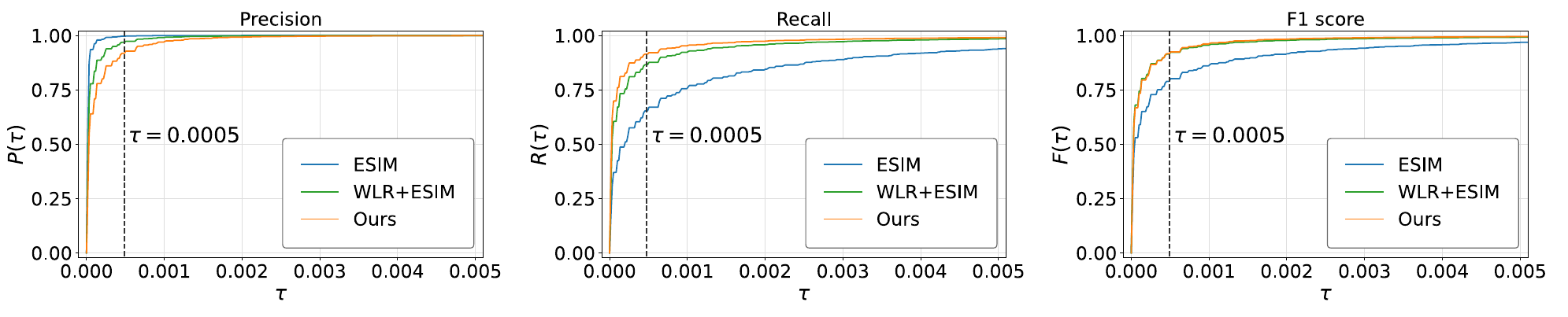}
  \vspace{-5mm}
  \caption{Comparison of precision, recall, and F1 scores of event detection. The distance threshold (i.e., $\tau = \num{5.0e-4}$) used to calculate the values in \cref{tab:quant-comparison} is depicted with a broken vertical line. Best viewed on screen.}
  \label{fig:metrics}
\end{figure*}

Our system consists of two components, i.e., the MC path tracing engine and image-to-event converter. The path-tracing engine is implemented on the top of PBRT-v4, an open-source path tracer~\cite{pbrt-v4}. As with the original WLR method~\cite{moon2014adaptive}, we obtain noisy frames with 9D pixel features, i.e., 2D image coordinate, 3D normal vector, 1D depth value, and 3D texture color at the first bounce. Our image-to-event converter to obtain event data is implemented with Python. While showing only the results for the San Miguel scene due to space limitation, we provide the results for other scenes and further analyses at \projpage.

The performance of our method is compared against two baseline approaches. The first baseline, ``ESIM'' applies ESIM~\cite{rebecq2018esim} directly to input noisy video frames\footnote{While ESIM is a method to obtain events at adaptive time points, the one here obtains those at time points with the fixed interval.}, while the second baseline ``WLR+ESIM'' applies ESIM after applying the WLR-based denoising~\cite{moon2014adaptive} to each video frame. We prepared noisy video frames obtained from 32, 64, and 128 path samples per pixel (spp) to test baseline methods and ours. In contrast, the reference data is obtained by ESIM applied to approximately noise-free video frames obtained from 4096 spp. Rendering a 240-frame video took about 20 minutes with 32 spp, while it took 20 hours with 4096 spp.

\heading{Visual comparison}
\Cref{fig:results} shows the visual comparison of event-based videos at three selected frames. As shown, directly applying ESIM to noisy video frames results in detecting many wrong events in the dark region (i.e., the inside of the building at the right) of the San Miguel scene. In contrast, WLR+ESIM detects events more appropriately and succeeded in suppressing the noise events in the dark region. The results of the proposed method are approximately equivalent to WLR+ESIM, but its computational complexity is significantly lower because our method only calculates the WLR model at only about \SI{28}{\percent} of pixels. Specifically, our method spent approximately \num{3.0} minutes on the computer equipped with Intel Core i9-9900K CPU (8 cores) and \SI{32}{\giga\byte} of RAM, while WLR+ESIM spent \num{4.5} minutes.

\begin{table}[t!]
  \centering
  \caption{Quantitative comparison of the proposed method with baselines. In this table, precision $P(\tau)$, recall $R(\tau)$, and F1 score $F(\tau)$ are computed with $\tau = \num{5.0e-4}$, which is depicted with broken vertical lines in \cref{fig:metrics}.}
  \label{tab:quant-comparison}
  {\tablefontsize
    \begin{tblr}{
        colspec={
            Q[l]
            Q[c,si={table-format=3}]
            Q[c,si={table-format=1.3,round-mode=places,round-precision=3}]
            Q[c,si={table-format=1.3,round-mode=places,round-precision=3}]
            Q[c,si={table-format=1.3,round-mode=places,round-precision=3},LightOrange]
            Q[c,si={table-format=1.5,round-mode=places,round-precision=5},LightOrange]
          },
        colsep=2.2mm,
        rowsep=0.4mm,
      }
      \toprule
       & {{{spp}}} & {{{$P(\tau)\UA$}}} & {{{$R(\tau)\UA$}}} & {{{$F(\tau)\UA$}}} & {{{CD$\DA$}}} \\
      \cmidrule[r]{1-1} \cmidrule{2-2} \cmidrule[l]{3-5} \cmidrule[l]{6-7}
      ESIM & 32 & \BB 0.997 & 0.653 & 0.789 & 0.000598      \\
      & 64        & \BB 0.996          & 0.684              & 0.811              & 0.000547      \\
      & 128       & \BB 0.995          & 0.740              & 0.849              & 0.000448      \\
      \cmidrule[r]{1-1} \cmidrule{2-2} \cmidrule[l]{3-5} \cmidrule[l]{6-7}
      WLR+ESIM & 32 & 0.969 & 0.866 & 0.914 & 0.000235      \\
      & 64        & 0.964              & 0.906              & \BB 0.934          & \BB 0.000167  \\
      & 128       & 0.959              & 0.931              & \BB 0.944          & \BB 0.000135  \\
      \cmidrule[r]{1-1} \cmidrule{2-2} \cmidrule[l]{3-5} \cmidrule[l]{6-7}
      Ours & 32 & 0.918 & \BB 0.914 & \BB 0.916 & \BB 0.000206  \\
      & 64        & 0.908              & \BB 0.935          & 0.922              & 0.000177      \\
      & 128       & 0.899              & \BB 0.948          & 0.923              & 0.000166      \\
      \bottomrule
    \end{tblr}
  }
\end{table}

\heading{Quantitative comparison}
The accuracy of event detection is evaluated by handling each set of events as a ``signed'' point set, i.e., $(x, y, t)$ with the sign of event polarization $p$. To this end, we modified the definitions of precision and recall for the point sets~\cite{tankstemples} by considering the event polarization. Let $d_{\vb{e} \rightarrow \mathcal{S}}$ be the distance from an element $\mathbf{e}$ to its closest entry in a set $\mathcal{S}$, $\mathcal{S}^{p}$ be the subset of $\mathcal{S}$ consisting of events with polarization $p$, and $\tau$ be the distance threshold to determine whether two events with the same polarization are close or not. Then, the precision $P(\tau)$ and recall $R(\tau)$ are defined as
\begin{subequations}
  \begin{align}
    P(\tau) &= \frac{1}{|\mathcal{G}|} \textstyle\sum_{p \in \{ +, - \}} \sum_{\vb{g} \in \mathcal{G}^{p}} \qty[ d_{\vb{g} \rightarrow \mathcal{R}^{p}} < \tau ], \label{eq:precision} \\
    R(\tau) &= \frac{1}{|\mathcal{R}|} \textstyle\sum_{p \in \{ +, - \}} \sum_{\vb{r} \in \mathcal{R}^{p}} \qty[ d_{\vb{r} \rightarrow \mathcal{G}^{p}} < \tau ], \label{eq:recall}
  \end{align}
\end{subequations}
Based on these definitions, we can also define the F1 score $F(\tau)$, i.e., the harmonic average of $P(\tau)$ and $R(\tau)$. Moreover, by this interpretation of events as points in the spatiotemporal space, we can calculate the chamfer distance (CD) of two sets of points by modifying the definition in \cite{yew2020-RPMNet} to be aware of the polarization sign.

By this definition, we can draw curves of the F1 score by taking $\tau$ to the horizontal axis. We compare the curves for the baseline methods and ours in \cref{fig:metrics}. In this figure, the higher the curve of the F1 score is located, the more the detected events are similar to those of reference. As shown in this figure, the curve of our system is located above the curve of ESIM and sufficiently close to that of WLR+ESIM. In addition, \cref{tab:quant-comparison} shows the values of $P(\tau)$, $R(\tau)$ and $F(\tau)$ when $\tau$ is set to \num{5.0e-4}, indicating that our system can obtain as good F1 scores and chamfer distances as time-consuming WLR+ESIM. WLR+ESIM often detects erroneous events due to the inconsistency of WLR models of two consecutive frames. In contrast, our method can avoid such false events by using a consistent WLR model over consecutive frames until a new event is detected. As a result, our method achieves even better scores than WLR+ESIM as shown in \cref{tab:quant-comparison}.

\heading{Bandwidth optimization}
The original WLR method~\cite{moon2014adaptive} determines two bandwidth parameters, i.e., $h$ and $\mathbf{b}$ in \cref{eq:weight-sum}, using the bias and variance of MC integration at each pixel. However, in our experiment, optimizing these values did not improve the quality of output event-based videos significantly, while the input video frames got visually better. Unfortunately, optimizing these parameters requires additional computational costs for performing the SVD of several $D \times D$ matrices. Specifically, when we optimize both $h$ and $\mathbf{b}$ as the original WLR method does, about 8 times more computation time is required. Moreover, this optimization requires storing the logarithmic brightness and variances of feature vectors during rendering, which needs uncommon modification of rendering engines. Therefore, we concluded solving the simple WLR model in \cref{eq:minimization} using $h=1$ and $\mathbf{b}_j = 1$ for all $j$ is a good choice for event detection.

\section{Conclusion}
\label{sec:conclusion}

This paper presented an efficient method to obtain event-based videos from noisy video frames. Our method applied the WLR~\cite{moon2014adaptive} for detecting changes in brightness, while achieved reducing the number of times of solving the regression problems. For future work, we would like to take the characteristics of DVS sensors~\cite{hu2021v2e} more carefully in our regression-based brightness change detection. We are also interested in extending the proposed model with a non-linear regression~\cite{moon2016adaptive} and deep-learning-based denoising method~\cite{bako2017kernel}. Developing an efficient path-tracing system is another direction, where path samples are adaptively collected when an event can more likely occur.

\heading{Acknowledgment}
This study is jointly supported by JSPS KAKENHI (JP21K19799, JP21H05054, JP19H04137), JST FOREST (JPMJFR206I), and JST MIRAI (JPMJMI19B2).

\vfill\pagebreak

\section*{References}

\bibliographystyle{IEEEbib}
\bibliography{refs}


\ifsubmit
\else
  \onecolumn
\makeatletter
\def\maketitle{\par
  \begingroup
  \def\thefootnote{}
  \def\@makefnmark{\hbox
    {$^{\@thefnmark}$\hss}}
  \if@twocolumn
    \twocolumn[\@maketitle]
  \else \newpage
    \global\@topnum\z@ \@maketitle \fi\@thanks
  \endgroup
  \setcounter{footnote}{0}
  \let\maketitle\relax
  \let\@maketitle\relax
  \gdef\thefootnote{\arabic{footnote}}\gdef\@@savethanks{}%
  \gdef\@thanks{}\gdef\@author{}\gdef\@title{}\let\thanks\relax}

\def\@maketitle{\newpage
  \null
  \vskip 2em \begin{center}
    {\large \bf \@title \par} \vskip 1.0em {\large \lineskip .5em
        \begin{tabular}[t]{c}\@name \\ \@address
        \end{tabular}\par} \end{center}
  \par
  \vskip 1.0em}
\makeatother

\setcounter{figure}{0}
\setcounter{table}{0}
\setcounter{page}{1}
\renewcommand{\thefigure}{A\arabic{figure}}
\renewcommand{\thetable}{A\arabic{table}}

\title{---Supplementary Document---\\[2mm]Event-based Camera Simulation using Monte Carlo Path Tracing\\with Adaptive Denoising}
\name{Yuta Tsuji$^1$ ~~~ Tatsuya Yatagawa$^{2}$ ~~~ Hiroyuki Kubo$^3$ ~~~ Shigeo Morishima$^1$}
\address{$^1$Waseda University ~~~ $^2$Hitotsubashi University ~~~ $^3$Chiba University}

\maketitle

\begin{table}[h!]
  \centering
  \begin{tblr}{
    colspec={X[c]X[c]},
    row{1} = {font=\Large\bfseries\sffamily}
    }
    ~~~~\textbf{Living room} & ~~~\textbf{Two boxes} \\
    \includegraphics*[width=\linewidth]{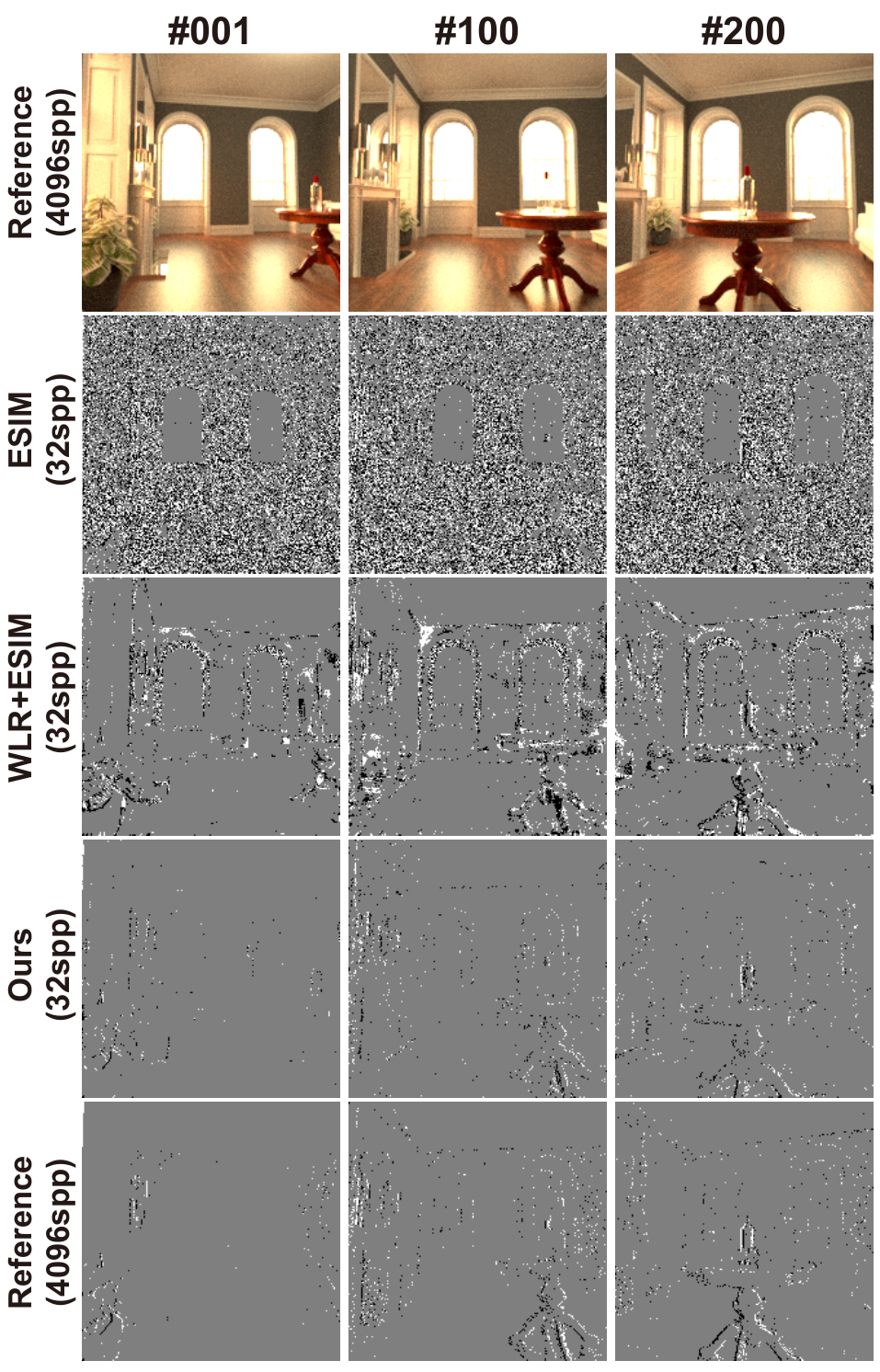} &
    \includegraphics*[width=\linewidth]{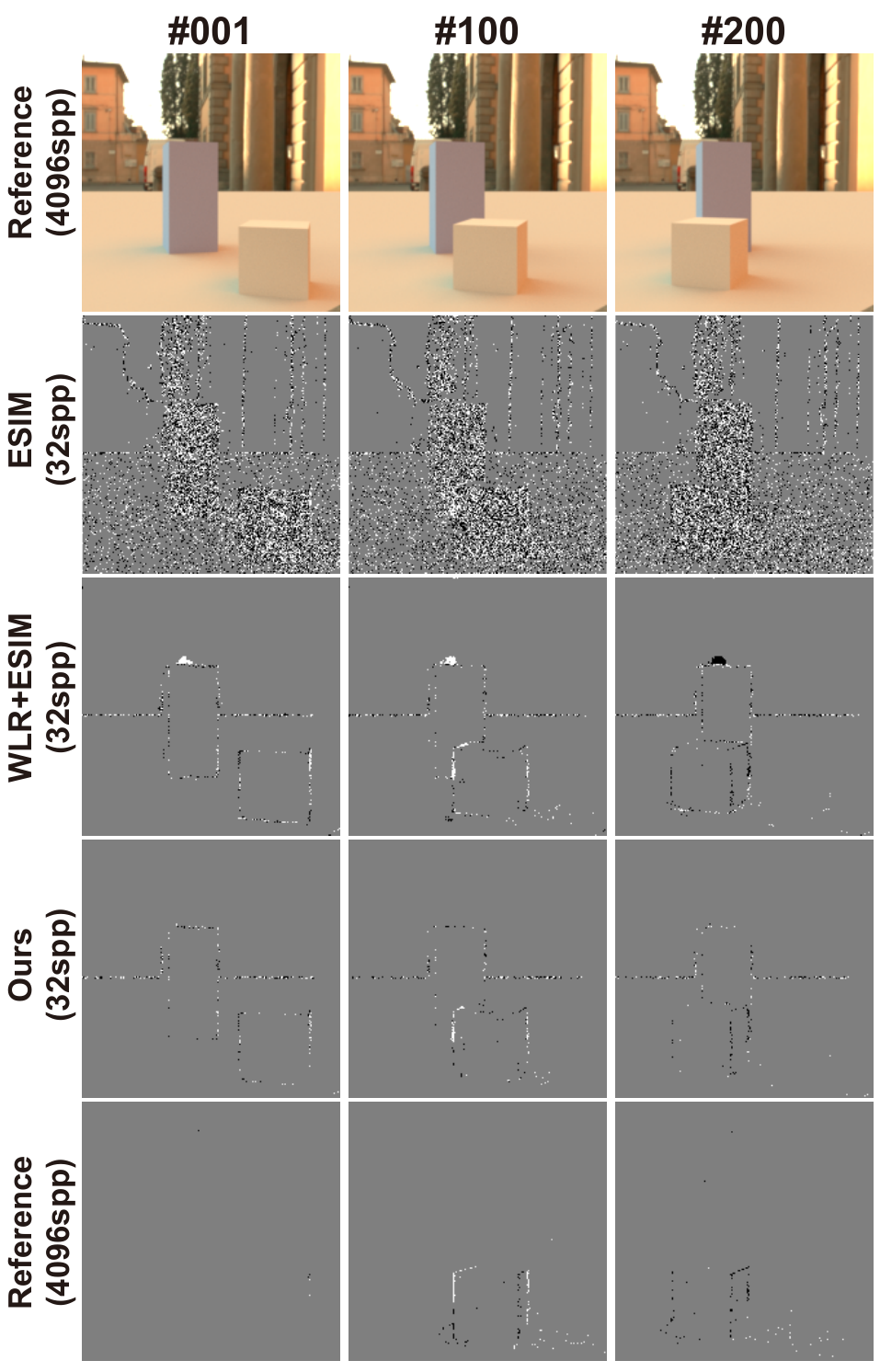}
  \end{tblr}
  \captionof{figure}{Visual comparison of event-based videos for ``Living room'' (left) and ``Two boxes'' (right) scenes. The 1st, 100th, and 200th frames are shown from left to right. The full-length videos for these results are available on our project page: \projpage. Although the results for the ``Living room'' scene reveal that our method is robust to the temporal incoherency of noisy pixel values, the WLR+ESIM significantly suffers from that problem. In contrast, both our method and WLR+ESIM have a limitation in that they tend to detect wrong events due to antialiasing at the objects' occluding contours (see the results for the ``Two boxes'' scene).}
  \label{fig:event-results-supl}
\end{table}

\begin{table}[t!]
  \centering
  \begin{tblr}{
    colspec={X[c]},
    rowsep=0mm,
    row{1,3} = {font=\large\bfseries\sffamily}
      }
    \textbf{Living room}                                     \\ \includegraphics*[width=\linewidth]{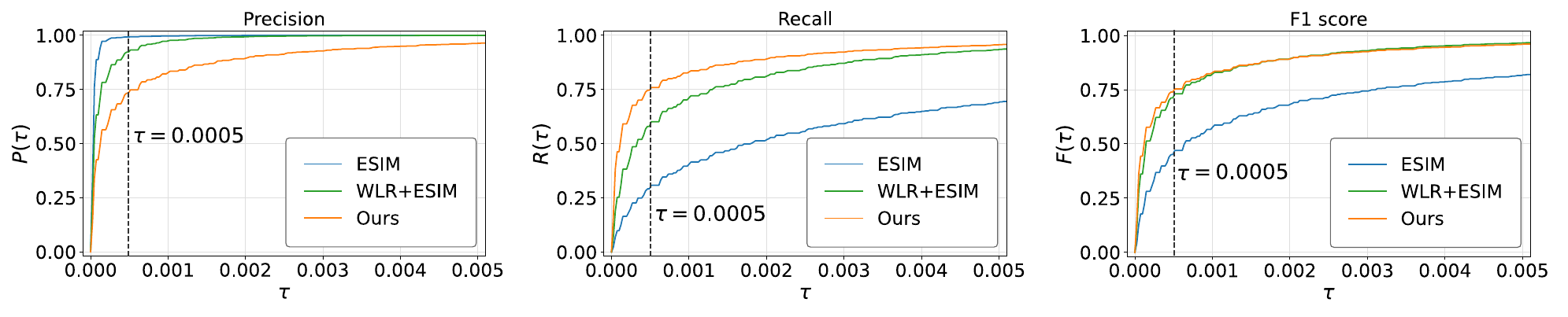} \\
    \textbf{Two boxes}                                                                                                       \\
    \includegraphics*[width=\linewidth]{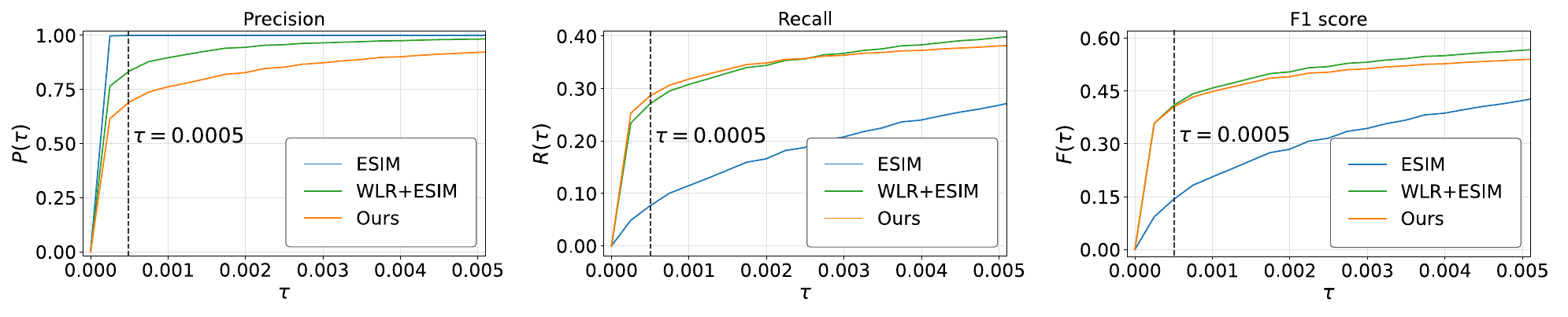}
  \end{tblr}
  \captionof{figure}{Comparison of precision, recall, and F1 scores of event detection. Those for the ``Living room'' scene are shown at the top, and those for the``Two boxes'' scene is shown at the bottom. Be careful that the scales for horizontal axes of the chart for ``Two boxes'' are adjusted to fit their ranges. As we explained in the caption of \cref{fig:event-results-supl}, the problem of detecting wrong events due to antialiasing decrease the recall and F1 scores for the ``Two boxes'' scene, even though these scores are approximately the same between our method and WLR+ESIM.}
  \label{fig:metrics-supl}
\end{table}

\begin{table}[t!]
  \centering
  \caption{Quantitative comparison of the proposed method with baselines using ``Living room'' (left) and ``Two boxes'' scenes. As \cref{tab:quant-comp-supl} in the main text, precision $P(\tau)$, recall $R(\tau)$, and F1 score $F(\tau)$ are computed with $\tau = \num{5.0e-4}$, which is depicted with broken vertical lines in \cref{fig:metrics-supl}.}
  \label{tab:quant-comp-supl}
  {\tablefontsize
    \begin{tblr}{
        colspec={
            Q[l]
            Q[c,si={table-format=3}]
            Q[c,si={table-format=1.3,round-mode=places,round-precision=3}]
            Q[c,si={table-format=1.3,round-mode=places,round-precision=3}]
            Q[c,si={table-format=1.3,round-mode=places,round-precision=3},LightOrange]
            Q[c,si={table-format=1.5,round-mode=places,round-precision=5},LightOrange]
            Q[l]
            Q[c,si={table-format=3}]
            Q[c,si={table-format=1.3,round-mode=places,round-precision=3}]
            Q[c,si={table-format=1.3,round-mode=places,round-precision=3}]
            Q[c,si={table-format=1.3,round-mode=places,round-precision=3},LightOrange]
            Q[c,si={table-format=1.5,round-mode=places,round-precision=5},LightOrange]
          },
        colsep=2.5mm,
        rowsep=1mm,
      }
      \cmidrule[0.3mm]{1-6}       \cmidrule[l,0.3mm]{7-12}
      \SetCell[c=6]{c,font=\large\bfseries\sffamily}Living room &  &  &  &  &  & \SetCell[c=6]{c,font=\large\bfseries\sffamily} Two boxes &  &  &  &  & \\
      \cmidrule[0.3mm]{1-6}       \cmidrule[l,0.3mm]{7-12}
      & {{{spp}}} & {{{$P(\tau)\UA$}}} & {{{$R(\tau)\UA$}}} & {{{$F(\tau)\UA$}}} & {{{CD$\DA$}}} &  & {{{spp}}} & {{{$P(\tau)\UA$}}} & {{{$R(\tau)\UA$}}} & {{{$F(\tau)\UA$}}} & {{{CD$\DA$}}} \\
      \cmidrule[r]{1-1} \cmidrule{2-2} \cmidrule[l]{3-6} \cmidrule[lr]{7-7} \cmidrule{8-8} \cmidrule[l]{9-12}
      ESIM & 32 & 0.99253 & 0.29118 & 0.45026 & 0.002984 & ESIM & 32 & 0.99973 & 0.07654 & 0.14220 & 0.033390      \\
      & 64        & 0.99029            & 0.33244            & 0.49777            & 0.002657      &                                                          & 64        & 0.99621            & 0.09275            & 0.16969            & 0.032088      \\
      & 128       & 0.98623            & 0.40691            & 0.57611            & 0.002080      &                                                          & 128       & 0.96628            & 0.12189            & 0.21647            & 0.031886      \\
      \cmidrule[r]{1-1} \cmidrule{2-2} \cmidrule[l]{3-6} \cmidrule[lr]{7-7} \cmidrule{8-8} \cmidrule[l]{9-12}
      WLR+ESIM & 32 & 0.92203 & 0.57952 & 0.71171 & 0.000796 & WLR+ESIM & 32 & 0.83474 & 0.27110 & 0.40928 & 0.021196      \\
      & 64        & 0.88844            & 0.63106            & 0.73795            & 0.000667      &                                                          & 64        & 0.83463            & 0.35862            & 0.50168            & 0.018829      \\
      & 128       & 0.84880            & 0.67354            & 0.75108            & 0.000633      &                                                          & 128       & 0.83065            & 0.47504            & 0.60442            & 0.015347      \\
      \cmidrule[r]{1-1} \cmidrule{2-2} \cmidrule[l]{3-6} \cmidrule[lr]{7-7} \cmidrule{8-8} \cmidrule[l]{9-12}
      Ours & 32 & 0.73280 & 0.74579 & 0.73924 & 0.000891 & Ours & 32 & 0.69134 & 0.28607 & 0.40469 & 0.020928      \\
      & 64        & 0.70719            & 0.80943            & 0.75487            & 0.000762      &                                                          & 64        & 0.66044            & 0.40091            & 0.49894            & 0.017521      \\
      & 128       & 0.67801            & 0.83918            & 0.75004            & 0.000758      &                                                          & 128       & 0.61094            & 0.52840            & 0.56668            & 0.014506      \\
      \cmidrule[0.3mm]{1-6}       \cmidrule[l,0.3mm]{7-12}
    \end{tblr}
  }
\end{table}

\begin{table}[t!]
  \centering
  \caption{Analysis of the effect for event detection threshold $C$. As this table shows, the event detection accuracy of the proposed method becomes higher than WLR+ESIM as the threshold $C$ is smaller. In contrast, when the threshold $C$ is larger, the event detection accuracy becomes slightly worse than WLR+ESIM, while the sparse event detection allows the proposed method to significantly shorten the computation time. Thus, our method has a tradeoff between the event detection accuracy and the computation time when the event detection threshold $C$ changes. The results here are computed for the video frames of the ``San Miguel'' scene rendered with 32spp. Note that the results in the main text are for $C=0.60$.}
  \label{tab:effect-threshold}
  {\tablefontsize
    \begin{tblr}{
        colspec={
            Q[l]
            Q[c,si={table-format=1.2}]
            Q[c,si={table-format=1.3,round-mode=places,round-precision=3}]
            Q[c,si={table-format=1.3,round-mode=places,round-precision=3}]
            Q[c,si={table-format=1.3,round-mode=places,round-precision=3},LightOrange]
            Q[c,si={table-format=1.5,round-mode=places,round-precision=5},LightOrange]
          },
        rowsep=0.4mm,
        colsep=3mm,
      }
      \toprule
       & {{{$C$}}} & {{{$P(\tau)\UA$}}} & {{{$R(\tau)\UA$}}} & {{{$F(\tau)\UA$}}} & {{{CD$\DA$}}} \\
      \cmidrule[r]{1-1} \cmidrule{2-2} \cmidrule[l]{3-5} \cmidrule[l]{6-7}
      ESIM & 0.25 & \BB 0.9989 & 0.85174 & 0.91989 & 0.000149      \\
      & 0.50      & \BB 0.99810        & 0.69719            & 0.82094            & 0.000446      \\
      & 0.75      & \BB 0.99405        & 0.60226            & 0.75007            & 0.000894      \\
      & 1.00      & \BB 0.98858        & 0.53833            & 0.69707            & 0.001576      \\
      \cmidrule[r]{1-1} \cmidrule{2-2} \cmidrule[l]{3-5} \cmidrule[l]{6-7}
      WLR+ESIM & 0.25 & 0.99380 & 0.91686 & 0.95378 & 0.000105      \\
      & 0.50      & 0.97714            & 0.87485            & 0.92317            & 0.000202      \\
      & 0.75      & 0.95327            & 0.85214            & \BB 0.89987        & \BB 0.000290  \\
      & 1.00      & 0.92683            & 0.82997            & \BB 0.87573        & \BB 0.000416  \\
      \cmidrule[r]{1-1} \cmidrule{2-2} \cmidrule[l]{3-5} \cmidrule[l]{6-7}
      Ours & 0.25 & 0.98743 & \BB 0.94463 & \BB 0.96556 & \BB 0.000086  \\
      & 0.50      & 0.94363            & \BB 0.92091        & \BB 0.93213        & \BB 0.000170  \\
      & 0.75      & 0.86118            & \BB 0.89825        & 0.87933            & 0.000292      \\
      & 1.00      & 0.76068            & \BB 0.86497        & 0.80948            & 0.000510      \\
      \bottomrule
    \end{tblr}
  }
\end{table}

\vspace{2em}

\begin{figure}[t!]
  \centering
  \includegraphics[width=0.85\linewidth]{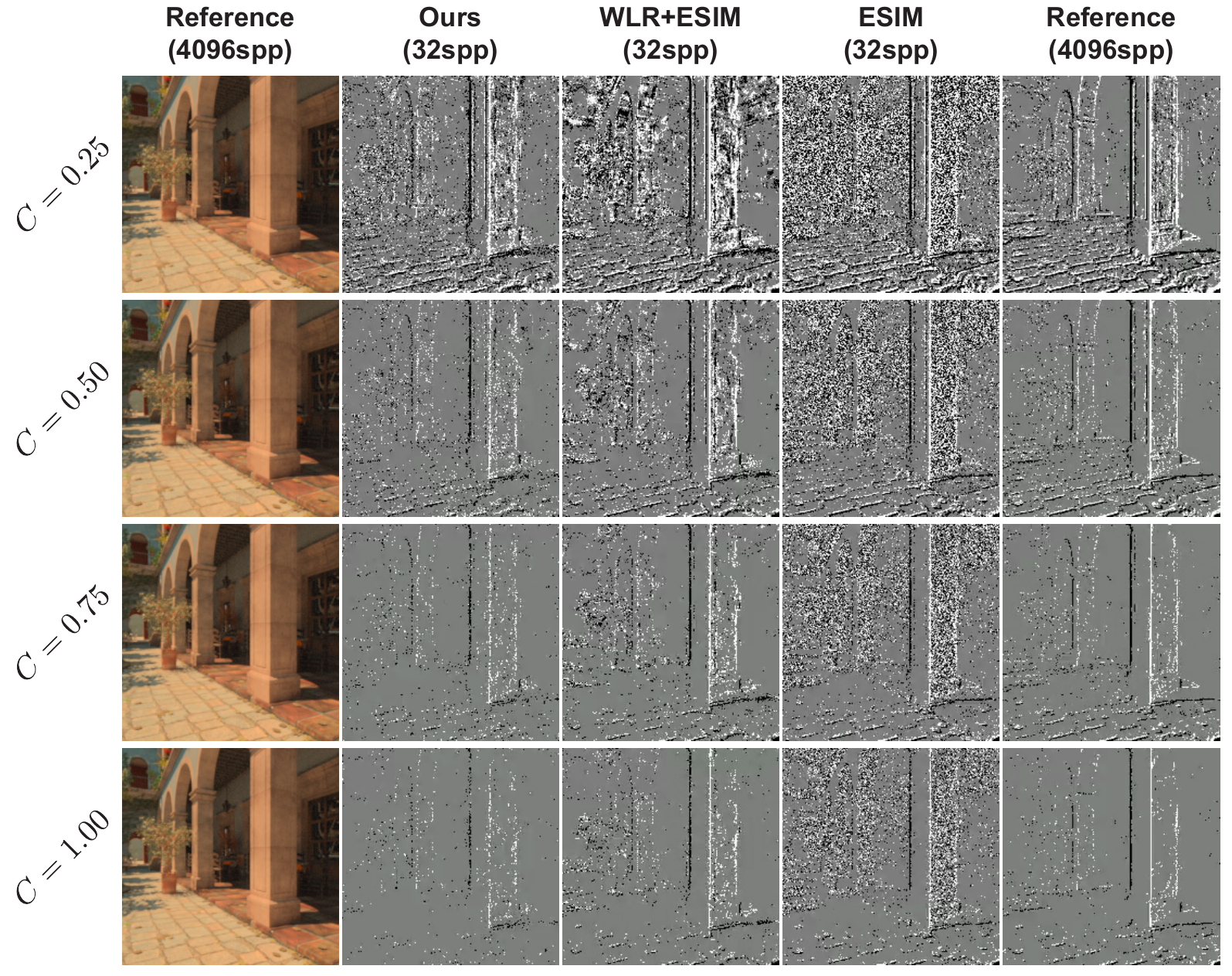}
  \caption{The effect of event detection threshold $C$ for the resulting event-based video (100th frame). As shown in this figure, our method obtains the results that are most similar to those of reference, while WLR+ESIM suffers from dappled artifacts due to the inconsistency of the WLR models of successive frames. The corresponding evaluation scores for these results are shown in \cref{tab:effect-threshold}.}
\end{figure}

\vfill\pagebreak

\fi

\end{document}